\documentclass[letterpaper, 10 pt, conference]{ieeeconf}  

\IEEEoverridecommandlockouts                              

\usepackage{cite}
\usepackage{amsmath,amssymb,amsfonts}
\usepackage{algorithmic}
\usepackage{graphicx}
\usepackage{textcomp}
\usepackage{xcolor}
\usepackage{algorithm}
\usepackage{verbatim} 
\usepackage{mathrsfs}
\usepackage{subfigure}
\usepackage{empheq}
\usepackage{lipsum}
\usepackage{comment}
\usepackage{mathtools, cuted}
\usepackage{bm}
\usepackage[utf8]{inputenc}
\usepackage[english]{babel}

\usepackage{amsthm}

\usepackage{soul}
\newcommand{\vg}[1]{\bm{#1}}
\renewcommand{\v}[1]{\mathbf{#1}}
\newtheorem{definition}{Definition}
\title{\LARGE \bf
Accelerated ADMM based Trajectory Optimization for Legged Locomotion with Coupled Rigid Body Dynamics}

\author{Ziyi Zhou$^{1}$ and Ye Zhao$^{2}$
\thanks{This work was supported in part by an NSF IIS Award No. 1924978.}
\thanks{The authors are with the Laboratory for Intelligent Decision and Autonomous Robots, George W. Woodruff School of Mechanical Engineering.}
\thanks{$^{1}$Ziyi Zhou is with School of Electrical and Computer Engineering, Georgia Institute of Technology, Atlanta, GA 30332, USA
        {\tt\small zhouziyi@gatech.edu}}%
\thanks{$^{2}$Ye Zhao is with George W. Woodruff School of Mechanical Engineering, Georgia Institute of Technology, Atlanta, GA 30313, USA
        {\tt\small ye.zhao@me.gatech.edu}}%
}

\begin{document}

\maketitle
\thispagestyle{empty}
\pagestyle{empty}

\begin{abstract}

Trajectory optimization is becoming increasingly powerful in addressing motion planning problems of underactuated robotic systems. Numerous prior studies solve such a class of large non-convex optimal control problems in a hierarchical fashion.
However, numerical accuracy issues are prone to occur when one uses a full-order model to track reference trajectories generated from a reduced-order model. This study investigates an approach of Alternating Direction Method of Multipliers (ADMM) and proposes a new splitting scheme for legged locomotion problems. Rigid body dynamics constraints and other general constraints such as box and cone constraints are decomposed to multiple sub-problems in a principled manner. The resulting multi-block ADMM framework enables us to leverage the efficiency of an unconstrained optimization method--Differential Dynamical Programming--to iteratively solve the optimizations using centroidal and whole-body models. Furthermore, we propose a Stage-wise Accelerated ADMM with over-relaxation and varying-penalty schemes to improve the overall convergence rate. We evaluate and validate the performance of the proposed ADMM algorithm on a car-parking example and a bipedal locomotion problem over rough terrains.
\end{abstract}





\section{INTRODUCTION}
Trajectory optimization approaches have been investigated extensively in the field of dynamic legged locomotion \cite{posa2014direct,tassa2014control, dai2014whole}. However, the optimal control problem of underactuated robots with high degrees of freedom (DoFs) commonly suffers from the curse of dimensionality, non-convexity, and intractable computation complexity. To address these issues, a majority of existing results employ a hierarchical approach: firstly solve the locomotion problem based on a reduced-order model such as linear inverted pendulum \cite{kajita20013d} or centroidal model \cite{orin2013centroidal}; then employ a whole-body model to track reference trajectories generated from the reduced-order model. Via this hierarchy, the complexity of optimization problems reduces significantly, and successful implementations have been achieved in \cite{dai2014whole}, \cite{herzog2015trajectory}. However, the mismatch between simplified and whole-body models often leads to severe tracking problems and even locomotion falls, given that the simplified model does not have a knowledge of how the angular momentum affect full-body motions.

Solving trajectory optimizations in a distributed and iterative manner provides an alternative direction to address such a mismatch across different models. The rationale of alternatively solving centroidal and whole-body models is initially proposed in \cite{herzog2016structured}. Recently, the work of \cite{budhiraja2019dynamics} leverages Alternating Direction Method of Multipliers (ADMM) to decompose the original optimal control of a locomotion problem into two sub-blocks iteratively, which minimizes their local cost functions. Multiple coupling constraints are introduced to achieve dynamics consensus between the centroidal and whole-body models. 
However, the above works share two common drawbacks: (i) the power of using a splitting scheme such as ADMM for locomotion dynamics decomposition has not been exploited to its full potential; (ii) the stopping criterion (or equivalently convergence criterion) is not clearly defined. Thus, the feasibility of the whole-body motions is not always guaranteed, although they claim that the algorithm can converge within a few iterations. In our study, we will analyze three ADMM variants and propose an accelerated method in a stage-wise fashion. A stopping criterion will be clearly defined to generate feasible whole-body motions.

Distributed optimization algorithms such as ADMM have numerous variants regarding specific applications and required convergence rates. A classical ADMM algorithm is a special case of Douglas-Rachford splitting methods \cite{eckstein1992douglas}, and is formulated in \cite{boyd2011distributed} for the first time to solve an optimization problem that is separable into two sub-problems with a consensus constraint. This algorithm guarantees a global convergence upon the premise that the sub-problems are closed, proper, and convex while the augmented Lagrangian has a saddle point (see \cite{boyd2011distributed}, Sec. III). A natural extension to this method is a Gauss-Seidel multi-block ADMM, which has been successfully evaluated in numerous real-world problems  \cite{peng2012rasl,tao2011recovering}. Although the general multi-block ADMM is not guaranteed to converge \cite{chen2016direct}, the study in \cite{lin2015sublinear} proves a sub-linear convergence rate under specific assumptions on convexity and a bounded Lagrangian multiplier coefficient.  Non-convex problems have also been studied in \cite{hong2016convergence}, where the convergence is merely guaranteed under specific conditions on objectives and penalty parameters. In this study, we aim at exploiting the distributed structure of this multi-block ADMM algorithm to solve legged locomotion problems possessing a well-established dynamics structure.

A salient feature of our proposed ADMM algorithm is to handle constraints in one designated sub-block while allowing other sub-blocks to solve unconstrained optimizations via off-the-shelf solvers such as Differential Dynamical Programming (DDP) or iterative Linear Quadratic Regulator (iLQR). Recently, advanced DDP-type optimization approaches have been proposed to handle constraints. In \cite{tassa2014control,xie2017differential}, an additional quadratic program is solved in each time step of the DDP backward pass. In \cite{howell2019altro}, an Augmented Lagrangian method is used to solve an unconstrained problem by converting constraints into penalty costs. Applying ADMM methods to convex quadratic cost functions with linear dynamics and box constraints is proposed in \cite{o2013splitting}. This work is extended to solve an optimization with cone constraints in \cite{o2016conic}. The study in \cite{sindhwani2017sequential} devises an outer loop based on Sequential Convex Programming (SCP) and applies consensus ADMM to handle constraints. Our strategy for addressing constraints is to take advantage of the ADMM distributed structure, and thoroughly decompose the whole-body dynamics. We design a new block for handling additional constraints without adjusting the same distributed framework. Compared to the constrained problem of each block solved by interior-point or active-set methods, the DDP-type operator splitting scheme employed in our study benefits from a low computational burden per ADMM iteration \cite{o2013splitting}. The reason lies in that unconstrained trajectory optimizations are solved for centroidal and whole-body models, respectively, leveraging the efficiency of DDP. 






In light of the discussions above, our contributions are summarized as follows: (i) proposing a novel distributed optimization framework to decompose the large-scale, highly complex optimization problem into multiple sub-problems, inspired by multi-block ADMM and an operator splitting method from \cite{o2013splitting,o2016conic}; (ii) devising a Stage-wise Accelerated ADMM (SWA-ADMM) with over-relaxation and varying-penalty schemes which improve the convergence rate than that of standard ADMM and its variants; 
(iii) providing a general, parallelizable framework capable of handling more dynamics and task constraints by introducing additional sub-blocks. This framework is applicable to general robotic systems involving Lagrangian rigid body and underactuated dynamics.

\section{Problem Formulation}
\label{sec:formulation}
Free-floating rigid body dynamics are widely investigated for modeling underactuated legged locomotion \cite{featherstone2014rigid, sentis2010compliant,  righetti2011inverse}. Given a sequence of predefined footsteps, we aim to generate whole-body feasible motions for a bipedal walking robot. With ground contact forces and underactuated dynamics, the whole-body dynamics are expressed as:
\begin{equation}\label{eq:lagrangian}
    \v H(\v q) \v {\ddot q} + \v C(\v q, \v {\dot q})  = \v B \v u + \v J^T_c \vg \lambda  
\end{equation}
where $\v q \in \mathbb{R}^{n}$ is the generalized joint configuration for a robot; $\v u \in \mathbb{R}^{m}$ is the control input; $\vg \lambda$ is the contact force of the corresponding foot contact point. The matrix $\v H \in \mathbb{R}^{n\times n}$ is the inertia matrix, $\v C \in \mathbb{R}^{n}$ is the sum of gravitational, centrifugal and Coriolis forces, $\v B \in \mathbb{R}^{n \times m}$ is the selection matrix, $\v J_c$ is the stacked Jacobian matrix for foot contact points.

Exploring the structure of Lagrangian dynamics in Eq.~(\ref{eq:lagrangian}) enables a straightforward decomposition. The first 6 rows represent underactuated dynamics of the center-of-mass (CoM) state while the remaining ones represent actuated dynamics at the joint level. The decomposed dynamics are formulated as:
\begin{subequations}
    \begin{align}
    \label{eq:decom-lagrang-a} & m \v {\ddot c} = \sum\nolimits_j \vg \lambda_j + m \v g \\\label{eq:decom-lagrang-b}
	& \v I \vg {\ddot \theta} = \sum\nolimits_j (\v p_j - \v c) \times \vg \lambda_j\\\label{eq:decom-lagrang-c}
	& \Bar{\v H}(\v q) \v {\ddot q} + \Bar{\v C}(\v q, \v {\dot q})  = \v u + \Bar{\v J}^T_c \vg \lambda  
	\end{align}
\end{subequations}
where Eqs.~(\ref{eq:decom-lagrang-a}) and (\ref{eq:decom-lagrang-b}) define a centroidal momentum model. The center of mass position, orientation and moment of inertia are denoted by $\v c$, $\vg \theta$, and $\v I$. The $j^{\rm th}$ contact point is $\v p_j$. The quantities for actuated joints are denoted by $\Bar{\v H}$, $\Bar{\v C}$ and $\Bar{\v J}_c$. We define $\v s=(\v c^T, \vg \theta^T, \v q^T, \v {\dot c}^T, \vg {\dot \theta}^T, \v {\dot q}^T)^T$ as the generalized state. The state of centroidal model is expressed as $\v s_{\rm cen}=(\v c^T, \vg \theta^T, \v {\dot c}^T, \vg {\dot \theta}^T)^T$ and the one for whole-body is $\v s_{\rm wbd}=(\v q^T, \v {\dot q}^T)^T$. The control inputs for centroidal and whole-body models are contact force $\vg \lambda$ and torque input $\v u$, respectively. A sequence of state-control pairs represents a trajectory $\vg \phi$. We have $\vg \phi_{\rm wbd} \coloneqq (\v s_{\rm wbd}[1, \ldots, T], \v u[1, \ldots, T-1])$ and $\vg \phi_{\rm cen} \coloneqq (\v s_{\rm cen}[1, \ldots, T], \vg \lambda[1, \ldots, T-1])$ representing trajectories of whole-body and centroidal models. This dynamic decomposition has been investigated in \cite{herzog2016structured} and \cite{budhiraja2019dynamics} to facilitate solving the optimization problem. 

Based on dynamics decomposition for centroidal and whole-body models, we formulate a general optimal control problem:
\begin{subequations}\label{eq:problem_formulation}
\begin{align}\nonumber
    & \underset{\vg \phi_{\rm cen}, \vg \phi_{\rm wbd}}{\text{min}} && \sum_{i = 1}^{N} \Big[\mathcal{L}_{\rm wbd}(\v s_{\rm wbd}[i], \v u[i]) + \mathcal{L}_{\rm cen}(\v s_{\rm cen}[i], \vg \lambda[i])  \Big]\\\label{eq:ocp-b}
	& \text{subject to} && \v H(\v q) \v {\ddot q} + \v C(\v q, \v {\dot q})  = \v B \v u + \v J^T_c g_{\lambda}(\v q, \v {\dot q}, \v u)  \\\label{eq:ocp-c}
	&&& m \v {\ddot c} = \sum\nolimits_{j} \vg \lambda_j + m \v g \\\label{eq:ocp-d}
	&&& \v I \vg {\ddot \theta} = \sum\nolimits_j (\v p_j - \v c) \times \vg \lambda_j \\\label{eq:ocp-e}
	&&& \v c = {\rm CoM}(\v q) \\\label{eq:ocp-f}
	&&& \begin{pmatrix}
	m \dot{\v c} \\
	\v I \vg {\dot \theta} 
	\end{pmatrix} = \boldsymbol{A}_g(\v q) \dot{\v q} \\\label{eq:ocp-g}
	&&& 	\vg \lambda = g_{\lambda}(\v q, \v {\dot q}, \v u) \\\label{eq:ocp-h}
	&&& \v s_{\rm} \in \mathcal{S} , \ \v u_{\rm} \in \mathcal{U} \\\label{eq:ocp-i}
	&&&  g_{\lambda_j}(\v q, \v {\dot q}, \v u) \in \mathcal{F}_j, \;\; \forall j \in \mathcal{I}_{\rm contact}
\end{align}
\end{subequations}
where $\mathcal{L}_{\rm wbd}$ and $\mathcal{L}_{\rm cen}$ stand for local cost functions of whole-body and centroidal models. A mapping from the generalized state $\v q$ to center of mass state $\v c$ is expressed as $\rm CoM(\cdot)$. The centroidal momentum matrix is denoted as $\boldsymbol{A}_g$ \cite{orin2013centroidal}. The function for computing contact forces of the whole-body model is $g_{\lambda}(\v q, \v {\dot q}, \v u)$ \cite{budhiraja2018differential}. We use $\mathcal{I}_{\rm contact} \coloneqq \{{\rm left, right} \}$ to index contact phases. 


Eqs.~(\ref{eq:ocp-b}), (\ref{eq:ocp-c}) and (\ref{eq:ocp-d}) are aforementioned dynamics equations. Eqs.~(\ref{eq:ocp-e}), (\ref{eq:ocp-f}) and (\ref{eq:ocp-g}) are named as dynamics consensus constraints to enforce consensus between whole-body and centroidal models for center of mass state, linear and angular momentum, and contact forces. Eqs.~(\ref{eq:ocp-h}) and (\ref{eq:ocp-i}) are additional constraints incorporated in this optimization problem. This formulation is analogous to the one in \cite{budhiraja2019dynamics} but solved using different ADMM algorithms introduced in next sections. As a minor comment, in \cite{budhiraja2019dynamics}, the friction cone constraint $\vg \lambda_j \in \mathcal{F}_j$ is incorporated in the centroidal model sub-block while we solve Eq.~(\ref{eq:ocp-i}) in the whole-body model. 

\section{Preliminaries}\label{sec:prelim}
In this section, we introduce the basic optimization formulations of two Alternating Direction Method of Multipliers (ADMM) variations.

\subsection{Multi-block ADMM}
ADMM algorithms solve optimization problems in a distributed manner. The classical ADMM is formulated in \cite{boyd2011distributed} to solve an optimization problem that is separable into two blocks with a linear coupling constraint. The optimization problem can be written as:
\begin{equation}
\begin{aligned}
    & \underset{\v x, \v z}{\text{min}} \ f(\v x) + g(\v z) \quad \text{s.t.} \ \boldsymbol{A}\v x+\boldsymbol{B}\v z=\v c
\end{aligned}
\end{equation}
where $\v x$ and $\v z$ are two sets of variables that construct a separable objective.

Multi-block ADMM is a natural extension to a more general case, where the objective function is separable into $N$ blocks ($N \ge 3$):
\begin{equation}
\begin{aligned}
    \underset{\v x_1,\v x_2, \ldots, \v x_N}{\text{min}} \ \sum_{i = 1}^{N} f_i(\v x_i) \quad  \text{s.t.} \ \sum_{i = 1}^{N}\boldsymbol{A}_i \v x_i=\v b
\end{aligned}
\end{equation}
where $\v x_i \in \mathbb{R}^{n_i}$, $\boldsymbol{A}_i \in \mathbb{R}^{m \times n_i}$, $\v b \in \mathbb{R}^m$.

The augmented Lagrangian is expressed as:
\begin{equation}\label{eq:augmented-larangian}
    \begin{aligned}
    L_{\rho}(\v x_1, \ldots, \v x_N, \v y)
    = \sum_{i = 1}^{N} f_i(\v x_i)+\v y^{T}(\sum_{i = 1}^{N}\boldsymbol{A}_i \v x_i-\v b)\\
    + \frac{\rho}{2}\|\sum_{i = 1}^{N}\boldsymbol{A}_i \v x_i-\v b\|^2
    \end{aligned}
\end{equation}
where $\v y$ is a dual variable and $\rho$ is an augmented Lagrangian parameter. The constant term in Eq.~(\ref{eq:augmented-larangian}) can be ignored to derive a simplified augmented Lagrangian using a scaled dual variable:
\begin{equation}\nonumber
    L_{\rho}(\v x_1, \ldots, \v x_N, \v w)=\sum_{i = 1}^{N} f_i(\v x_i)+\frac{\rho}{2}\|\sum_{i = 1}^{N}\boldsymbol{A}_i \v x_i-\v b+\v w\|^2
\end{equation}
where $\v w_{k}=\v y_{k}/\rho$. The scaled form is compact and easier to work with \cite{boyd2011distributed}. Similar to the standard ADMM, this multi-block ADMM contains a sequential update for both primal and dual variables:
\begin{subequations}
    \begin{align}\label{eq:multi-admm a}
    & \v x_1^{k+1}:=\underset{\v x_1}{\arg\min} \ L_{\rho}(\v x_1,\v x_2^k, \ldots,\v x_N^k; \v w^k) \\\nonumber
    & \ldots \\ \nonumber
    & \v x_i^{k+1}:=\underset{\v x_i}{\arg\min} \ L_{\rho}(\v x_1^{k+1}, \ldots , \v x_{i-1}^{k+1}, \v x_i, \v x_{i+1}^{k}, \ldots, \v x_N^k; \v w^k) \\\label{eq:multi-admm b}
    & \ldots \\ \label{eq:multi-admm c}
    & \v x_N^{k+1}:=\underset{\v x_N}{\arg\min} \ L_{\rho}(\v x_1^{k+1}, \ldots, \v x_{N-1}^{k+1}, \v x_N; \v w^k) \\\label{eq:multi-admm d}
    & \v w^{k+1} :=\v w^{k}+(\sum_{i = 1}^{N}A_i \v x_i^{k+1}-\v b)
\end{align}
\end{subequations}

\subsection{Consensus ADMM}
The consensus ADMM is a special case of multi-block ADMM and solves the following optimization:
\begin{equation}\label{eq:consensusADMM}
\begin{aligned}
    \underset{\Bar{\v x},\v x_1,\v x_2, \ldots, \v x_N}{\text{min}} \ \sum_{i = 1}^{N} f_i(\v x_i)+g(\Bar{\v x}) \\
    \text{s.t.} \ \v x_i=\Bar{\v x} \quad i=1,\dots,N
\end{aligned}
\end{equation}
where $\Bar{\v x}$ is a global set of decision variables and $g$ is a regularization function such as an indicator function to penalize constraint violations of state decision variables. Note that this type of consistency constraint can also be applied to control decision variables. This indicator function regarding a closed convex set $C$ is defined as:
\begin{equation}
    I_C(\v x,\v u)=\begin{cases}
    0, \ (\v x,\v u)\in C \\
    +\infty, \ \text{otherwise}
    \end{cases}
\end{equation}
where infinite values enforce convex constraints on states and controls.

\section{Distributed Trajectory Optimization}\label{sec:algorithm}
Multi-block ADMM is suitable for splitting the locomotion problem in Sec.~\ref{sec:formulation} into multiple blocks. In this section, we will analyze and implement three ADMM variants and compare their convergence performance. In particular, we propose a novel accelerated multi-block ADMM method and demonstrate the faster convergence rate of the proposed methods.

\subsection{Operator splitting of locomotion dynamics}
%

\begin{figure}[t]
    \centering
    \includegraphics[width=3in]{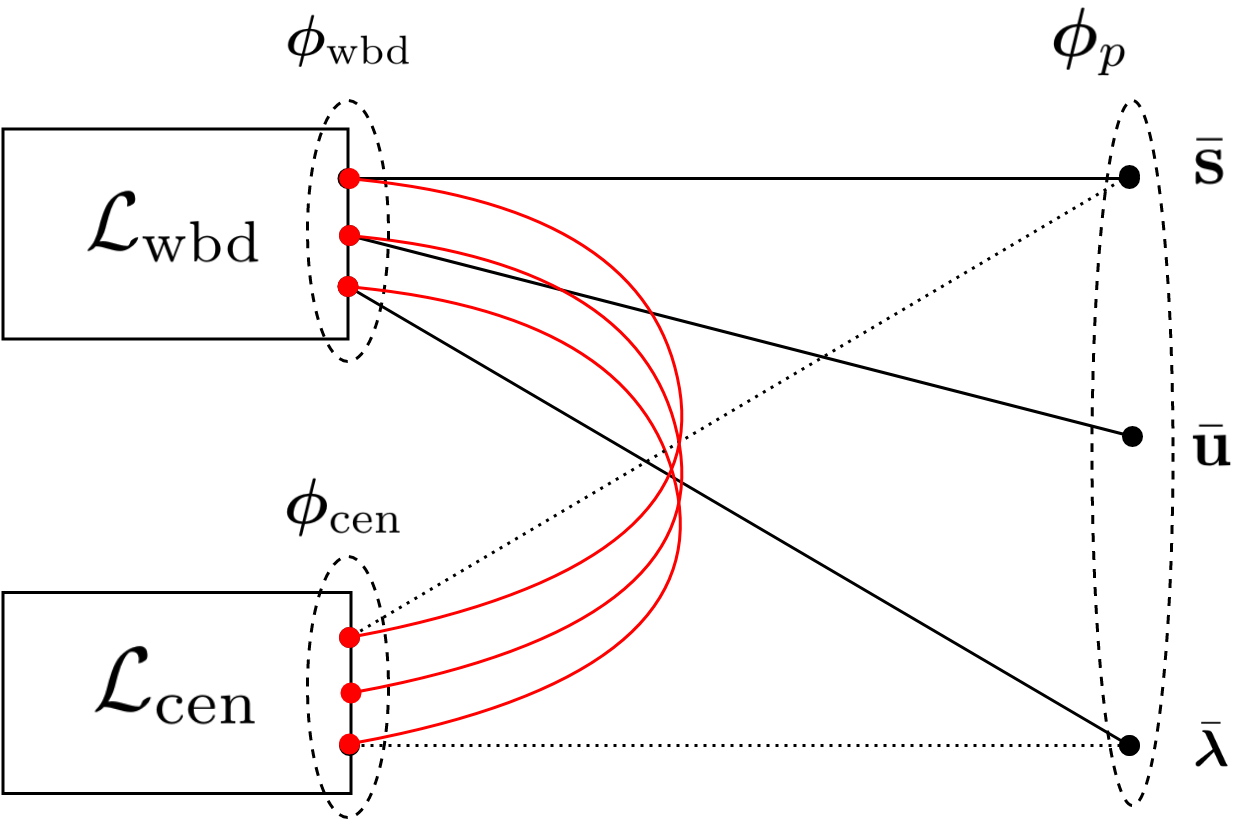}
    \caption{Structure of the splitting problem in Eqs.~(\ref{eq:three-block}) and (\ref{eq:coupling constraints}). The left illustrates local objective terms for whole-body and centroidal models. $\phi_{\rm wbd}$, $\phi_{\rm cen}$ and $\phi_{\rm p}$ are variable trajectories of the whole-body, centroidal and projection sub-blocks respectively. The nodes in this graph denote decision variables from each sub-block. Each black edge refers to a consistency constraint while a red edge represents a nonlinear dynamics consensus constraint. An edge with a dash line denotes another consistency constraint but not implemented in our study.} 
    \label{fig:graph}
    \vspace{-0.1in}
\end{figure}

 By defining a set of state and control variable copies, we use an indicator function and induced consistency constraints to incorporate bounding inequality constraints in Eqs.~(\ref{eq:ocp-h}) and (\ref{eq:ocp-i}). Similar to the decision variables of centroidal and whole-body sub-blocks, the newly-introduced decision variables are denoted as $\vg \phi_{p}=(\Bar{\v s}[1, \ldots, T], \Bar{\v u}[1, \ldots, T-1], \Bar{\vg \lambda}[1, \ldots, T-1])$, where the subscription $p$ denotes a projection on certain admissible sets. By adding the third sub-block, the optimal control problem in (\ref{eq:problem_formulation}) is reformulated as:
\begin{align}\nonumber
        \underset{\vg \phi_{\rm cen}, \vg \phi_{\rm wbd},\vg \phi_p}{\text{min}} \sum_{i = 1}^{N} \Big[& \mathcal{L}_{\rm wbd}(\v s_{\rm wbd}[i], \v u[i])
        + \mathcal{L}_{\rm cen}(\v s_{\rm cen}[i], \vg \lambda[i])\\\label{eq:three-block}
        & + I_{\mathcal{S},\mathcal{U},\mathcal{F}}(\Bar{\v s}[i], \Bar{\v u}[i], \Bar{\vg \lambda}[i]) \Big]
\end{align}
where $I_{\mathcal{S},\mathcal{U},\mathcal{F}}$ is the indicator function of joint-limit, torque-limit and friction cone constraint in Eqs.~(\ref{eq:ocp-h}) and (\ref{eq:ocp-i}). The coupling constraints in this study are defined as:
\begin{definition}[Coupling constraints]\label{definition:coupling}
Coupling constraints in our multi-block framework are equality constraints connecting decision variables from two distinct sub-blocks.
\begin{equation}\label{eq:coupling constraints}
\begin{aligned}
&\begin{rcases}
    {\rm CoM}(\v q) = \v c \\
    \boldsymbol{A}_g(\v q)\Dot{\v q}=
    \begin{bmatrix}
        m\Dot{\v c}\\\
        \v I \Dot{\vg \theta}
    \end{bmatrix}\\
    g_{\lambda}(\v q,\Dot{\v q},\v u)=\vg \lambda \\
    \end{rcases} \text{dynamics consensus constraints} 
    \end{aligned}
\end{equation}
\begin{equation}\nonumber
\begin{aligned}
&\begin{rcases}
    \v s = \Bar{\v s} \\
    \v u = \Bar{\v u} \\
    g_{\lambda}(\v q,\Dot{\v q},\v u) = \Bar{\vg \lambda}
    \end{rcases} 
    \text{projection consistency constraints}
\end{aligned}
\end{equation}
\end{definition}
%
\begin{definition}[Consistency constraints]
Consistency constraints are the last three constraints of coupling constraints in Definition~\ref{definition:coupling} and make consensus between decision variables from two distinct sub-blocks.
\end{definition}
The first three dynamics consensus constraints are analogous to the work of Budhiraja et.al \cite{budhiraja2019dynamics}. Distinct from their splitting method incorporating the friction cone constraint into the centroidal model sub-block, three more consistency constraints are introduced in our study to form a projection sub-block. As shown in Fig.~\ref{fig:graph}, the local cost functions of centroidal and whole-body sub-blocks and global projection copies form a bipartite graph \cite{boyd2011distributed}. Each edge of this graph denotes a coupling constraint between two sets of variables\footnote{For other underactuated robotic systems, we design different (i) variables for global projection copies and (ii) edges between local and global variables according to specific constraints in Eqs.~(\ref{eq:ocp-h}) and (\ref{eq:ocp-i})}. The consistency constraints between local variables and global projection copies ensure the former ones are inside a desired convex set. For instance, $\v s=\Bar{\v s}$ corresponds to an indicator function $I_{\mathcal{S}}$, ensuring $\v s$ to stay within a convex set $S$ in Eq.~(\ref{eq:ocp-h}). Note that the friction cone constraint in Eq.~(\ref{eq:ocp-i}) and contact force consensus constraint in Eq.~(\ref{eq:ocp-g}) are redundant in our locomotion problem. Therefore we only define a projection from $g_{\lambda}(\v q, \dot{\v q}, \v u)$ to a global variable copy $\Bar{\vg \lambda}$. 

Given the formulation above, we design a decomposition method analogous to multi-block ADMM. Each of the coupling constraints in Eq.~(\ref{eq:coupling constraints}) has an augmented Lagrangian parameter $\rho_i$ and a corresponding dual variable $\v w_i$ with $\ i \in \mathcal{I} \coloneqq \{c,h,\lambda,j,t,f\}$. We define a set of primal residuals $\v r_i$ for all the constraints in Eq.~(\ref{eq:coupling constraints}):
\begin{equation}
    \begin{aligned}[c]
        & \v r_c={\rm CoM}(\v q)-\v c \\
        & \v r_h=\boldsymbol{A}_g(\v q)\Dot{\v q}-
        \begin{bmatrix}
            m\Dot{\v c}\\\
            \v I \Dot{\vg \theta}
        \end{bmatrix}\\
        & \v r_{\lambda}=g_{\lambda}(\v q,\Dot{\v q},\v u)-\vg \lambda\\
        \end{aligned}
        \quad \quad \quad
        \begin{aligned}[c]
        & \v r_j=\v s-\Bar{\v s}\\
        & \v r_t=\v u-\Bar{\v u}\\
        & \v r_f=g_{\lambda}(\v q,\Dot{\v q},\v u)-\Bar{\vg \lambda}
    \end{aligned}
\end{equation}

The following steps solve each sub-problem and update the dual variables iteratively in a scaling form.
\begin{subequations}\label{eq:updates}
    \begin{align}
        & \nonumber \text{Primal updates:} \\\label{eq:primal-a}
        & \vg \phi_{\rm cen}^{k+1}:=\underset{\vg \phi_{\rm cen}}{\arg\min} \ \sum_{i = 1}^{N} \mathcal{L}_{\rm cen}(\vg \phi_{\rm cen})+ \sum_{i=c,h,\lambda}\frac{\rho_i}{2}\|\v r_i+\v w_i^k\|_2^2 \\\label{eq:primal-b}
        & \vg \phi_{\rm wbd}^{k+1}:=\underset{\vg \phi_{\rm wbd}}{\arg\min} \ \sum_{i = 1}^{N} \mathcal{L}_{\rm wbd}(\vg \phi_{\rm wbd}) + \sum_{i\in\mathcal{I}}\frac{\rho_i}{2}\|\v r_i+\v w_i^k\|_2^2\\\label{eq:primal-c}
        & \vg \phi_{p}^{k+1}:=\underset{\vg \phi_p}{\arg\min} \ \sum_{i = 1}^{N} I_{\mathcal{S},\mathcal{U},\mathcal{F}}(\vg \phi_{p}) + \sum_{i=j,t,f}\frac{\rho_i}{2}\|\v r_i+\v w_i^k\|_2^2
        \end{align}
\end{subequations}
\begin{subequations}
    \begin{align}\nonumber
         & \text{Dual updates:} \\
        & \v w_c^{k+1}=\v w_c^{k}+{\rm CoM}(\v q^{k+1})-\v c^{k+1} \tag{13d}\\
        & \v w_h^{k+1}=\v w_h^{k}+\boldsymbol{A}_g(\v q^{k+1})\Dot{\v q}^{k+1}-
        \begin{bmatrix}
            m\Dot{\v c}^{k+1}\\\
            \v I \Dot{\vg \theta}^{k+1}
        \end{bmatrix}\tag{13e}\\
        & \v w_{\lambda}^{k+1}=\v w_{\lambda}^{k}+g_{\lambda}(\v q^{k+1},\Dot{\v q}^{k+1},\v u^{k+1})-\vg \lambda^{k+1} \tag{13f}\\
        & \v w_j^{k+1}=\v w_j^{k}+\v s^{k+1}-\Bar{\v s}^{k+1} \tag{13g}\\
        & \v w_t^{k+1}=\v w_t^{k}+\v u^{k+1}-\Bar{\v u}^{k+1} \tag{13h}\\
        & \v w_f^{k+1}=\v w_f^{k}+g_{\lambda}(\v q^{k+1},\Dot{\v q}^{k+1},\v  u^{k+1})-\Bar{\vg \lambda}^{k+1} \tag{13i}
    \end{align}
\end{subequations}
where the elements $\v r_i$ in Eqs.~(\ref{eq:primal-a})-(\ref{eq:primal-c}) are continuously recomputed with updated decision variables over iterations. Detailed expressions of $\v r_i$ are ignored due to limited space. Eqs.~(\ref{eq:primal-a}) and (\ref{eq:primal-b}) are only subject to centroidal and whole-body dynamics respectively without any other additional constraints.

In this new operator splitting scheme, each sub-block of the centroidal and whole-body models is an unconstrained optimization problem and are both solved by DDP. Note that in Eq.~(\ref{eq:primal-b}), the contact constraint is incorporated in the forward pass to obtain contact-consistent ground reaction forces, which is similar to the approach in \cite{budhiraja2018differential}. The whole-body contact force $g_{\lambda}(\v q, \v {\dot q}, \v u)$ is expressed as new state variable. To cope with the constraints $\v s \in \mathcal{S},\v u \in \mathcal{U}$ and $\vg \lambda \in  \mathcal{F}$, the sub-block in Eq.~(\ref{eq:primal-c}) reduces to a projection operator on the admissible sets $\mathcal{S}$, $\mathcal{U}$ and $\mathcal{F}$: 
\begin{equation}
    \begin{aligned}\nonumber
         \vg \phi_{p}^{k+1}& :=\underset{\vg \phi_p \in C}{\arg\min} \ \sum_{i=j,t,f}\frac{\rho_i}{2}\|\v r_i+\v w_i\|_2^2\\
         C&:=\{(\Bar{\v s},\Bar{\v u},\Bar{\vg \lambda})|\Bar{\v s} \in \mathcal{S}, \Bar{\v u} \in \mathcal{U}, \Bar{\vg \lambda} \in  \mathcal{F}\}
    \end{aligned}
\end{equation}
which is interpreted as a simple saturation function. This projection operator is capable of handling cone constraints \cite{boyd2011distributed}. This subspace projection can be implemented more efficiently (see \cite{o2016conic} Sec.~4). 


The whole iterative process is illustrated in Algorithm 1. Decision variables $\vg \phi_{\rm cen}^0$ and $\vg \phi_{\rm wbd}^0$ are warm-started to make each sub-block mutually solvable and the output from previous ADMM iteration is used as a warm-start for the current ADMM iteration. 


\begin{algorithm}[t]
  \caption{Multi-block ADMM solver for locomotion}
  \begin{algorithmic}[1]
   \STATE $\vg \phi_{\rm cen} \gets \vg \phi_{\rm cen}^{0},\vg \phi_{\rm wbd} \gets \vg \phi_{\rm wbd}^{0},\vg \phi_{p} \gets \vg \phi_{p}^{0}$
   
   \STATE $\v w_i \gets \v w_i^0,\rho_i \gets \rho_i^0, i=\mathcal{I}$
   
    \REPEAT
    \STATE $\vg \phi_{\rm cen} \gets \text{DDP-centroidal}(\vg \phi_{\rm cen},\vg \phi_{\rm wbd},\vg \phi_{p},\v w_{i}, \rho_i),\ i=\{c,h,\lambda\}$
    
        \STATE $\vg \phi_{\rm wbd}\gets \text{DDP-wholebody}(\vg \phi_{\rm wbd},\vg \phi_{\rm cen},\vg \phi_{p},\v w_{i},\rho_i),\ i=\mathcal{I}$
        
        \STATE $\vg \phi_{p}\gets \text{Projection}(\vg \phi_{\rm wbd},\vg \phi_{\rm cen},\v w_i,\rho_i) \quad i=\{j,t,f\}$
        
        \STATE $\v w_i \gets \v w_i + \v r_i \quad i=\mathcal{I}$
    \UNTIL{stopping criterion is satisfied}
  \end{algorithmic}
\end{algorithm}

\subsection{Convergence analysis and stopping criteria}
The convergence of our algorithm is not formally guaranteed due to highly nonlinear dynamics constraints and nonlinear coupling constraints in Eq.~(\ref{eq:coupling constraints}). However, it is realized that if dynamics consensus constraints are satisfied, our multi-block operator splitting method reduces to a consensus ADMM, whose convergence is guaranteed under mild conditions \cite{hong2016convergence}. In practice, multi-block operator splitting methods are powerful and have a broad spectrum of applications for large-scale non-convex optimization problems.

To determine when the convergence is achieved, a stopping criterion is designed for all the coupling constraints. We use the $l_2$-norm of primal residuals with a tolerance $\epsilon$:
\begin{equation}\nonumber
    \|\v r_i^k\|_2 \leq \epsilon_i^{\rm pri}, \ i = \mathcal{I}
\end{equation}
where $\epsilon^{\rm pri}$ are predefined positive scalars.

In practice, achieving high accuracy of all constraint residuals is challenging \textit{yet} unnecessary. For instance, constraints related to walking robot contact forces often have large values due to robot gravity. In that case, relatively large residuals make these constraints difficult to satisfy the small tolerance values in stopping criteria but do not significantly affect the result accuracy. In this study, we suggest specifying the primal tolerance to be on a similar order of magnitude as the physical variable in each constraint. Meanwhile, a maximum iteration is set up as one stopping criterion. If the tolerances are relatively small, the ADMM sometimes terminates by hitting the maximum iteration to avoids more iteration runs without further cost reductions.

On the other hand, the local penalty costs sometimes do not converge to appropriate tolerance values that guarantee feasible whole-body motions. Although it is not easy to explicitly assess the feasibility of generated whole-body motions, we design an additional stopping criterion for the local cost reduction to remedy the shortcoming of using residual-only stopping criteria:
\begin{equation}
    |\mathcal{L}_{\rm wbd}^{k}-\mathcal{L}_{\rm wbd}^{k-1}| \leq \epsilon^{\rm cost}
\end{equation}
where the local cost tolerance $\epsilon^{\rm cost}$ plays a similar role as to the cost reduction in DDP.

With predefined tolerances $\epsilon^{\rm pri}$ and $\epsilon^{\rm cost}$, feasible motions are generated for complex locomotion behaviors such as walking over rough terrain. This type of locomotion behaviors usually requires more iterations to converge in the context of the standard ADMM set-up.

\subsection{ADMM variations for improving convergence rate}
The overall operator splitting method can generate solutions with moderate accuracy after the first few iterations and solve large-scale problems effectively \cite{o2013splitting}. However, a major defect of standard ADMM algorithms is its sub-linear convergence rate leading to unsatisfactory accuracy. To address this hurdle, we aim at an accelerated variant of the ADMM method to achieve a faster convergence rate.

In our locomotion formulation, since the ADMM framework we are using in Eq.~(\ref{eq:updates}) has nonlinear coupling constraints among multiple sub-blocks, standard accelerated methods with linear constraints are not directly applicable. In addition, dynamics constraints in each sub-block are highly nonlinear, resulting in a highly non-convex optimization problem. 

In this section, we first introduce multiple ADMM variations that improve the convergence rate. Based on these variations, we study three modified ADMM methods, compare their performance, and advocate an accelerated version, which numerically improves the convergence rate of consistency constraints in our multi-block operator splitting method. 

\subsubsection{Over-relaxation}
For a two-block ADMM,  over-relaxation is to introduce a new parameter $\alpha$ and replace $\boldsymbol{A}_1 \v x_1^{k+1}$ in Eq.~(\ref{eq:multi-admm b}) by: 
\begin{equation}\label{eq:relaxation}
    \alpha \boldsymbol{A}_1 \v x_1^{k+1} - (1-\alpha) (\boldsymbol{A}_2 \v x_2^{k}-\v b)
\end{equation}
where $\alpha \in (0,2)$ is a relaxation parameter. In particular, $\alpha<1$ denotes under-relaxation, while $\alpha>1$ denotes over-relaxation. This scheme has been studied in \cite{eckstein1992douglas}, and experiments have empirically shown that values of $\alpha \in [1.5,1.8]$ improve the convergence. 

\subsubsection{Varying-penalty parameter}
Another variation is to adjust the penalty parameter $\rho$ per iteration according to the relative values of primal and dual residuals \cite{boyd2011distributed}. As such, the convergence performance will be less sensitive to the initial value of $\rho$. This varying-penalty parameter is defined as:
\begin{equation}\label{eq:varying-penalty}
    \rho^{k+1}:= \begin{cases}
    \tau^{\rm incr}\rho^{k} \quad & \text{if} \ \|\v r^k\|_2^2 > \mu \|\v d^k\|_2^2\\
    \rho^{k}/ \tau^{\rm decr} \quad &  \text{if} \ \|\v d^k\|_2^2 > \mu \|\v r^k\|_2^2 \\
    \rho^{k} \quad & \text{otherwise}
    \end{cases}
\end{equation}
where $\mu>1$, $\tau^{\rm incr}>1$, and $\tau^{\rm decr}>1$. The dual residual $\v d^k$ for projection constraints are expressed by:
\begin{equation}
\begin{aligned}
    & \v d_j^{k}=\rho_j^{k}(\Bar{\v s}^k-\Bar{\v s}^{k-1}),\
     \v d_t^{k}=\rho_t^{k}(\Bar{\v u}^k-\Bar{\v u}^{k-1}),\\
    & \v d_f^{k}=\rho_f^{k}(\Bar{\vg \lambda}^{k}-\Bar{\vg \lambda}^{k-1})
    \end{aligned}
\end{equation}
Typically, the choices could be $\mu=10$ and $\tau^{\rm incr}=\tau^{\rm decr}=2$. The essence of adapting penalty parameters in each iteration is interpreted as maintaining the primal and dual residuals within a factor $\mu$ of one another as they converge to zero.


\subsubsection{Fast ADMM}
Goldstein et al. \cite{goldstein2014fast} proposed a Fast ADMM for the two-block problem based on a Nestorov acceleration method \cite{nesterov1983method}. The Fast ADMM has been proved to guarantee a quadratic convergence rate under strong convexity assumptions for each sub-block. A predictor-corrector step is applied to achieve this acceleration, where new sets of primal and dual variables are updated sequentially with a varying coefficient (More details are provided in \cite{goldstein2014fast}). 

However, the strong convexity assumption restricts this algorithm to be only applicable for highly structured optimization problems. Again, this algorithm does not apply to our locomotion problem which contains centroidal and whole-body non-linear dynamics. For the projection sub-block, it has a non-smooth and non-convex saturation function. 

Based on the ADMM variations aforementioned, we define a Stage-wise Accelerated ADMM (SWA-ADMM) as below. 
\begin{definition}[SWA-ADMM]
    Stage-wise Accelerated ADMM is defined as the multi-block ADMM based operator splitting method in Eq.~(\ref{eq:updates}), incorporating the over-relaxation in Eq.~(\ref{eq:relaxation}) and varying-penalty in Eq.~(\ref{eq:varying-penalty}) schemes in a stage-wise fashion to achieve an accelerated convergence rate.
\end{definition}
Convergence rate of the proposed SWA-ADMM outperforms that of the standard multi-block ADMM based operator splitting method in Eq.~(\ref{eq:updates}) under the condition that ADMM variations are implemented appropriately.
Empirically, applying the over-relaxation facilitates the convergence rate during the first few iterations \cite{o2013splitting}. Meanwhile, adopting a varying-penalty scheme enables us to achieve satisfactory small primal residuals. However, there is a caveat to apply them simultaneously. If the varying-penalty mechanism is applied from the first iteration, the scaling parameters in Eq.~(\ref{eq:varying-penalty}) possibly make the penalty cost coefficient $\rho$ become unreasonably large. Accordingly, the convergence of local cost functions $\mathcal{L}_{\rm wbd}$ and $\mathcal{L}_{\rm cen}$ deteriorates and requires more iterations.


With the caveat of applying over-relaxation and varying-penalty schemes elaborated above, we propose to integrate these two schemes into our multi-block framework in a stage-wise fashion. In Algorithm 2, we use over-relaxation from the beginning with $\alpha=1.65$. Then we apply the varying-penalty and over-relaxation schemes simultaneously after ${k}_{\rm sw}^{\rm th}$ iteration, such that the primal residual converges faster without over-penalizing constraint violations during the first few iterations. The performance of three ADMM variants will be compared in the next section.


\begin{algorithm}[t]
  \caption{Stage-wise Accelerated ADMM solver }
  \begin{algorithmic}[1]
  
    \STATE $\vg \phi_{\rm cen} \gets \vg \phi_{\rm cen}^{0},\vg \phi_{\rm wbd} \gets \vg \phi_{\rm wbd}^{0},\vg \phi_{p} \gets \vg \phi_{p}^{0}$
   
   \STATE $\v w_i \gets \v w_i^0,\rho_i \gets \rho_i^0,i=\mathcal{I}$
   
    \REPEAT
    \STATE $\vg \phi_{\rm cen} \gets \text{DDP-centroidal}(\vg \phi_{\rm cen},\vg \phi_{\rm wbd},\vg \phi_{p},\v w_{i}, \rho_i) \linebreak i=\{c,h,\lambda\} $
    
        \STATE $\vg \phi_{\rm wbd}\gets \text{DDP-wholebody}(\vg \phi_{\rm wbd},\vg \phi_{\rm cen},\vg \phi_{p},\v w_{i},\rho_i) \linebreak i=\mathcal{I}$
        
        \STATE $ \vg \phi_p^{\prime} \gets \alpha (\v s,\v u, \vg \lambda)+(1-\alpha)(\Bar{\v s},\Bar{\v u},\Bar{\vg \lambda})$
        
        \STATE $\vg \phi_{p}\gets \text{Projection}(\vg \phi_{p}^{\prime},\v w_i,\rho_i) \quad i=\{j,t,f\}$
        
        \STATE $\v w_i \gets \v w_i + \v r_i \quad i=\mathcal{I}$
        
        \IF{$\text{current iteration}>k_{\rm sw}$}
            \IF{$\|\v r_i\|_2^2 > \mu \|\v d_i\|_2^2$}
                \STATE $\rho_{i}=\tau^{\rm incr}\rho_{i}$
            \ELSIF{$\|\v d_i\|_2^2 > \mu \|\v r_i\|_2^2$}
                \STATE $\rho_{i}=\rho_{i}/ \tau^{\rm decr}$
            \ENDIF
        \ENDIF
        
    \UNTIL{stopping criterion is satisfied}
  \end{algorithmic}
\end{algorithm}


\section{Results}\label{sec:experiment}
In this section, we evaluate the multi-block ADMM and SWA-ADMM algorithms using a car-parking problem and a kneed compass gait walker. The locomotion example is implemented in a robotics optimization toolbox Drake \cite{drake}.

\subsection{Car-parking problem}
The car-parking problem with bounded control inputs is a well studied example in control-limited DDP \cite{tassa2014control} and TROSS \cite{sindhwani2017sequential}. This example enables us to benchmark the performance of the proposed ADMM solvers. The operator splitting for this system is only for a control-limit constraint, and therefore a two-block ADMM problem is studied.

The car dynamics have a 4-dimensional state vector, $\boldsymbol{\hat{x}}=(x,y,\theta,v)$ where $(x,y)$ is the car position, $\theta$ is the car angle relative to the x-axis, and $v$ is the velocity of front wheels. Control inputs are the front wheel angle $\omega$ and acceleration $a$, with limits of $\pm 0.5$ rad and $\pm 2.0$ ${\rm m/s^2}$, respectively. The objective is to move the car from an initial state $(1,1,\frac{3\pi}{2},0)$ to a parking goal state $(0,0,0,0)$. 

\begin{figure}[t]
    \centering
    \includegraphics[width=3.5in]{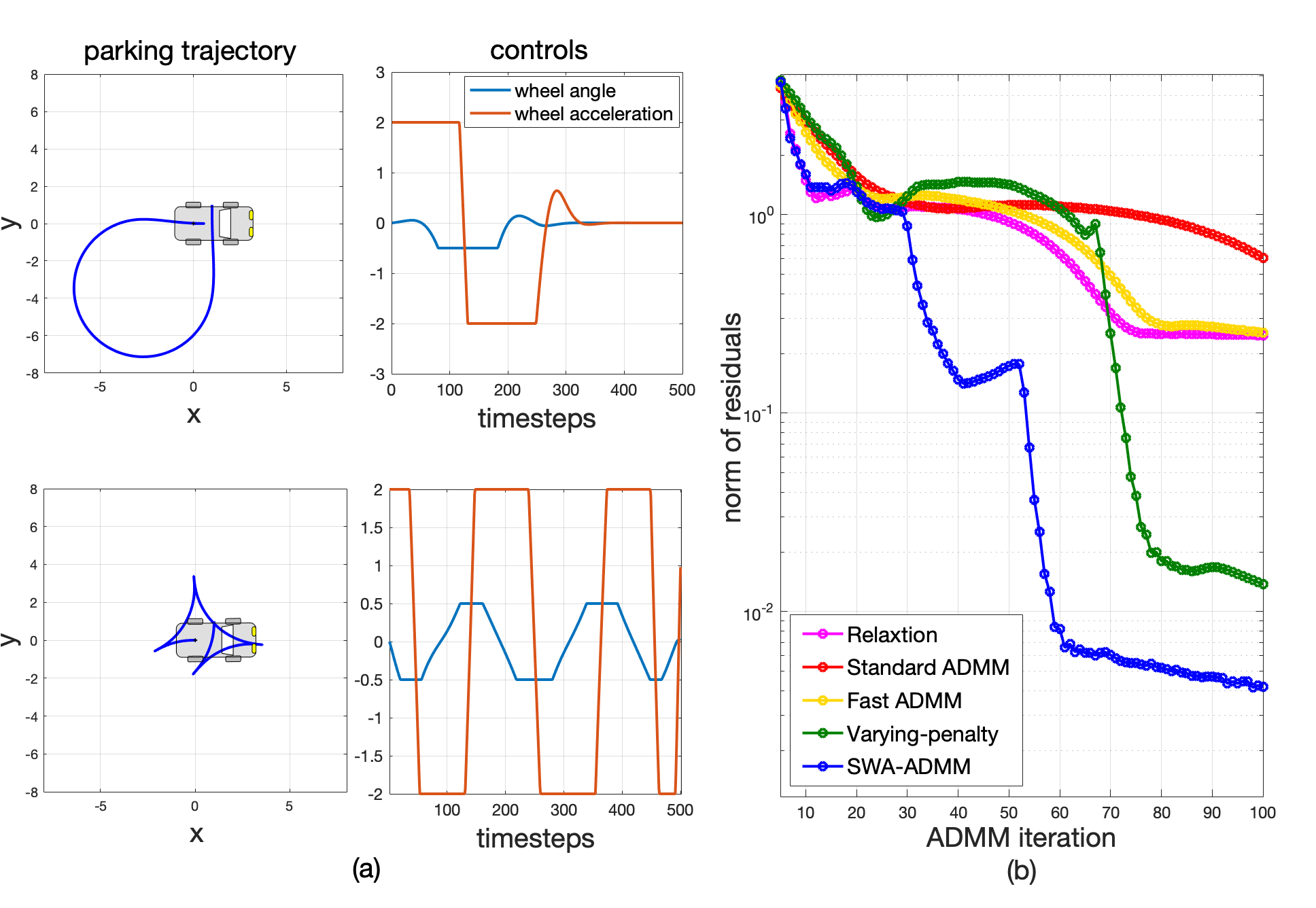}
    \vspace{-0.3in}
    \caption{(a) Comparison of parking trajectories and controls between our ADMM optimizer and control-limited DDP \cite{tassa2014control}. The upper two figures are our results while the bottom ones use control-limited DDP. The left column displays parking trajectories after convergence while the right column displays the control inputs. (b) Comparison of residuals between ADMM variants. The blue line shows the proposed SWA-ADMM method.}
    \label{fig:car_parking}
    \vspace{-0.15in}
\end{figure}

In this example, we apply the same running cost and final cost as the one in \cite{tassa2014control}. In SWA-ADMM set-up, the initial Lagrangian parameter $\rho$ is set as 0.01. Zero control inputs are used as an initial guess. Dual variables are initialized as zero. The varying-penalty parameters are set as $\mu=10$ and $\tau^{\rm incr}=\tau^{\rm decr}=2$ while $k_{\rm sw}$ is set as $16$. For the purpose of benchmarking, our DDP uses the same parameters as the ones in \cite{tassa2014control}. The resulting trajectories and the performances for multiple ADMM variants are shown in Fig.~\ref{fig:car_parking}.
%
%

%
Two observations are stated as below: (i) the generated car trajectory of our SWA-ADMM solver is significantly distinct from the one generated by control-limited DDP, indicating a different descent path for these two methods. Also, our method has a lower total cost after convergence than theirs. We attribute this observation to the different descent path induced by the ADMM algorithm; (ii) the SWA-ADMM shows a better convergence rate compared with other ADMM variants.  

\subsection{Locomotion problem}
A kneed compass gait walker is used to evaluate our proposed ADMM algorithms (see Fig.~\ref{fig:rough_walk}). This underactuated walker is comprised of two legs and has three actuators in total: one is a hip joint while the other two are knee joints. The operator splitting method via ADMM aims at dynamics consensus between reduced-order and whole-body models.

\subsubsection{Cost function design}
For the whole-body sub-block, we use (i) augmented Lagrangian penalty costs, (ii) quadratic costs to regularize torques and a final posture of the robot, (iii) soft constraints for foot trajectories. For the centroidal sub-block, contact forces and CoM velocities are regularized. More sophisticated terms can be incorporated within each sub-block as needed.

\subsubsection{Flat terrain walking}
\begin{figure*}[t]
    \centering
    \includegraphics[width=7in]{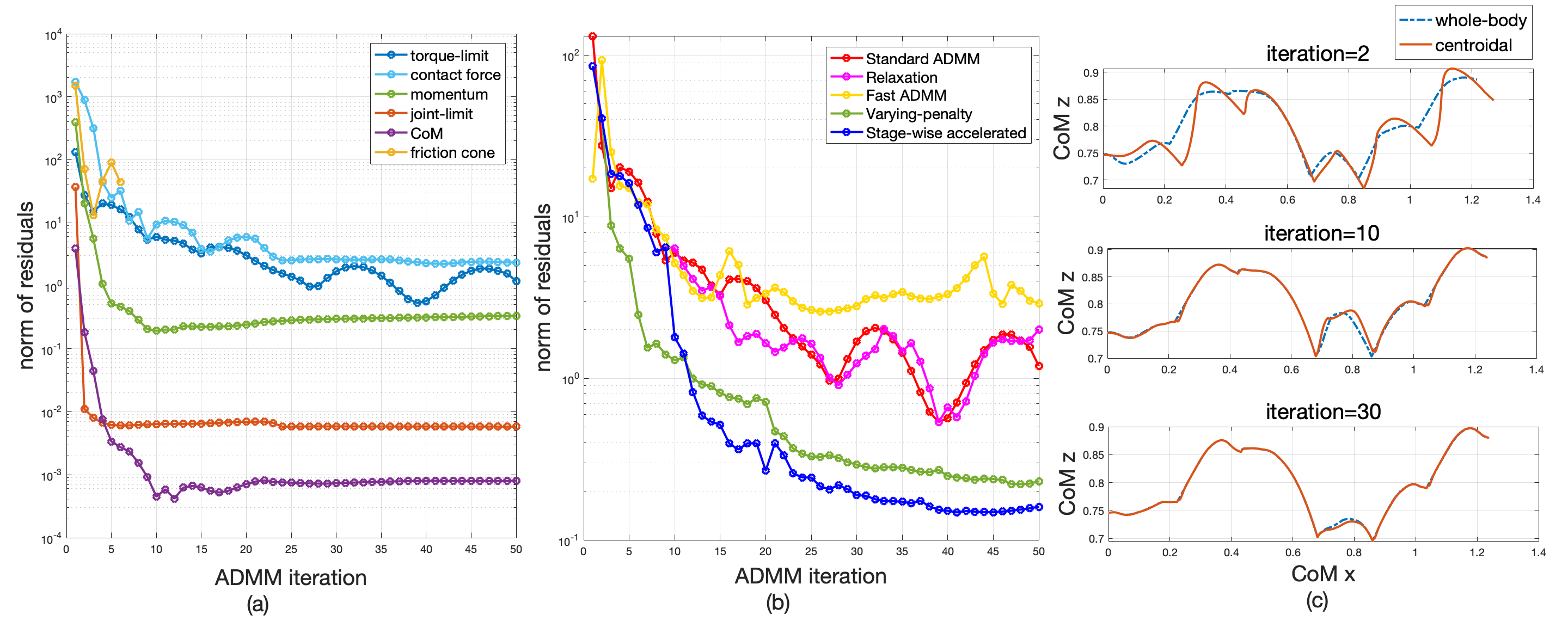}
    \vspace{-0.2in}
    \caption{(a) Residuals of the coupling constraints using multi-block ADMM operator splitting method. The contact force constraint has a relatively larger residual norm due to its compensation of the robot gravity. The residual norm of friction cone constraint after the first six iterations is not shown since it decreases to 0. (b) Residuals of torque-limit constraint in different ADMM variants for walking three steps on flat terrain. SWA-ADMM converges faster than other methods after 10 iterations. (c) CoM trajectories of walking 6 steps over rough terrain in the sagittal-vertical plane for $2^{\rm nd}, 10^{\rm th}$ and $30^{\rm th}$ iterations. }
    \label{fig:locomotion}
\end{figure*}

We study a scenario of multiple locomotion steps on flat ground. The ADMM optimization is solved in a model-predictive-control fashion with one walking step horizon. Time step is 0.01s and the horizon has $T = 50$ time steps. We incorporate all six constraints in Eq.~(\ref{eq:coupling constraints}) where the knee joint ranges from $[0,\pi]$ and the friction cone constraint has a friction coefficient 1. The augmented Lagrangian parameters are: $\rho_c=10^4,\rho_h=10^{-2},\rho_{\lambda}=10^{-2},\rho_j=10,\rho_t=0.1$ and $\rho_f=10^{-2}$. Fig.~\ref{fig:locomotion} (a) demonstrates residuals for all the coupling constraints. The maximum ADMM iteration is set as 50. After around 15 iterations for each walking step, all the residuals and local penalty costs converge to an acceptable accuracy. In this example, an evaluation on torque-limit constraint for different methods is shown in Fig.~\ref{fig:locomotion} (b), where our SWA-ADMM method outperforms other ADMM variants. The varying-penalty parameters are set as $\mu=10, \tau^{\rm incr}=\tau^{\rm decr}=5$ and the stage-wise iteration number $k_{\rm sw} = 9$.
The computation cost per ADMM iteration mainly comes from the DDP solver for whole-body model. After the first 2-3 ADMM iterations, the number of DDP iterations within each ADMM iteration only requires fewer than 10 to reach a cost reduction on the order of magnitude of $10^{-5}$.
\begin{figure}[t]
    \centering
    \includegraphics[width=2.3in]{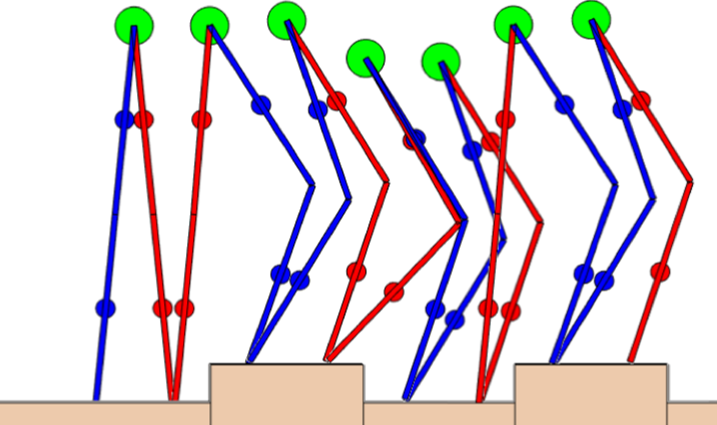}
    \caption{Walking sequence snapshot for rough terrain generated by the proposed SWA-ADMM solver.}
    \label{fig:rough_walk}
    \vspace{-0.15in}
\end{figure}



\subsubsection{Rough terrain walking}
We study a multiple-step locomotion behavior over rough terrain. The augmented Lagrangian parameters and stopping criteria are set to be the same as the ones for the flat terrain case. To illustrate the consensus process, the CoM trajectories of centroidal and whole-body models at $2^{\rm nd}, 10^{\rm th}$ and $30^{\rm th}$ ADMM iterations are shown in Fig.~\ref{fig:locomotion} (c). It is observed that after the first 2-3 ADMM iterations, the walker falls down, and the swing foot hits stairs. We realize that although the residuals decrease significantly, similar to the flat terrain case, the local penalty costs are still large, indicating motion infeasibility. After around 30 ADMM iterations for each walking step, the constraint violation and local cost reduction are all within the tolerances. A walking sequence on rough terrains is shown in Fig.~\ref{fig:rough_walk}. Note that fewer ADMM iterations are required if a better warm-start is applied. However, finding better warm-start sometimes is not straightforward, especially for locomotion over complex terrains.

One limitation of our SWA-ADMM method is that the acceleration performance depends on the selection of the initial augmented Lagrangian parameter $\rho$ for each projection consistency constraint. When the initial $\rho$ is less than 1, the SWA-ADMM method accelerates effectively. Overall, our SWA-ADMM algorithm is less sensitive to the initial parameter $\rho$ compared to the non-accelerated version.

\section{CONCLUSIONS}\label{sec:conclusions}
This study proposes a distributed, high-accuracy trajectory optimization for whole-body dynamic locomotion over rough terrain. We propose a general and efficient framework based on multi-block ADMM which splits the locomotion trajectory optimization problem into three sub-blocks. Each sub-block of centroidal and whole-body models solves an unconstrained optimization using DDP, while a third projection block handles box and cone constraints. To improve the convergence rate, a Stage-wise Accelerated ADMM is proposed via over-relaxation and varying-penalty schemes. A decent stopping criterion is designed to generate feasible whole-body motions. In our experiments, we validate the proposed splitting methods over flat and rough terrains.


\section*{ACKNOWLEDGEMENT}
The authors would like to express our special thanks to Rohan Budhiraja for his feedback and discussions at an early stage of this research.

\bibliographystyle{IEEEtran}
\bibliography{paper.bib}

\addtolength{\textheight}{-12cm}   



\end{document}